\documentclass[conference]{IEEEtran}
\IEEEoverridecommandlockouts
\usepackage{cite}
\usepackage{amsmath,amssymb,amsfonts}
\usepackage{algorithmic}
\usepackage{tikz}
\usepackage{graphicx}
\usepackage{textcomp}
\usepackage{xcolor}
\usepackage{booktabs}
\def\BibTeX{{\rm B\kern-.05em{\sc i\kern-.025em b}\kern-.08em
    T\kern-.1667em\lower.7ex\hbox{E}\kern-.125emX}}
\begin{document}

\title{Evaluating the Application of Large Language Models to Generate Feedback in Programming Education}

\author{\IEEEauthorblockN{1\textsuperscript{st} Sven Jacobs}
\IEEEauthorblockA{\textit{Computer Science Education} \\
\textit{University of Siegen}\\
Siegen, Germany \\
sven.jacobs@uni-siegen.de}
\and
\IEEEauthorblockN{2\textsuperscript{nd} Steffen Jaschke}
\IEEEauthorblockA{\textit{Computer Science Education} \\
\textit{University of Siegen}\\
Siegen, Germany \\
steffen.jaschke@uni-siegen.de}
}

\newcommand\copyrighttext{%
  \footnotesize \textcopyright \the\year{} IEEE. Personal use of this material is permitted. Permission from IEEE must be obtained for all other uses, in any current or future media, including reprinting/republishing this material for advertising or promotional purposes, creating new collective works, for resale or redistribution to servers or lists, or reuse of any copyrighted component of this work in other works.
  }
\newcommand\copyrightnotice{%
\begin{tikzpicture}[remember picture,overlay]
\node[anchor=south,yshift=10pt] at (current page.south) {\fbox{\parbox{\dimexpr0.75\textwidth-\fboxsep-\fboxrule\relax}{\copyrighttext}}};
\end{tikzpicture}%
}
\newcommand\submittedtext{%
  \footnotesize This work has been submitted to the IEEE for possible publication. Copyright may be transferred without notice, after which this version may no longer be accessible.}

\newcommand\submittednotice{%
\begin{tikzpicture}[remember picture,overlay]
\node[anchor=south,yshift=10pt] at (current page.south) {\fbox{\parbox{\dimexpr0.65\textwidth-\fboxsep-\fboxrule\relax}{\submittedtext}}};
\end{tikzpicture}%
}
\renewcommand\fbox{\fcolorbox{red}{white}}
\setlength{\fboxrule}{2pt} 

\maketitle
\copyrightnotice

\begin{abstract}
This study investigates the application of large language models, specifically GPT-4, to enhance programming education. The research outlines the design of a web application that uses GPT-4 to provide feedback on programming tasks, without giving away the solution. A web application for working on programming tasks was developed for the study and evaluated with 51 students over the course of one semester. The results show that most of the feedback generated by GPT-4 effectively addressed code errors. However, challenges with incorrect suggestions and hallucinated issues indicate the need for further improvements.
\end{abstract}

\section{Introduction}
In courses with numerous exercises, such as programming courses, providing feedback can be time-consuming for teachers. Simultaneously, it can be disadvantageous for learners if they have to wait too long for feedback before they can continue working on the assignment. To address this issue, many automated solutions have been developed in recent years \cite{jeuring.2022}. The development of large language models (LLMs) like GPT-4 \cite{openai.2023} has opened up a wide range of new possibilities in this area \cite{prather.2023} \cite{denny.2023a}. GPT-4's ability to solve introductory programming tasks is already around 95\% \cite{kiesler.2023b}. It can even solve and explain more difficult tasks effectively \cite{bubeck.2023}, and its performance can be further improved with prompting strategies \cite{nori.2023}. Google Deepmind's AlphaCode 2 demonstrates that the combination of fine-tuning and prompting strategies can already reach the 85th percentile on average on the code contest platform Codeforces \cite{alphacode.2023}.

While applications like ChatGPT or GitHub Copilot, which use LLMs, can help with programming, their primary purpose is to increase productivity rather than to aid learning. To be effective for skill development, such tools would require specific prompts to ensure that the correct solution is not provided immediately, allowing learners to engage in the problem-solving process. Therefore, we integrated GPT-4 into a new programming practice environment in an introductory computer science course to provide students with timely feedback. The research question for this work is: To what extent is the large language model GPT-4 able to provide feedback for programming education?

\section {Related work}
In the context of programming exercises, a variety of tools are used that already support different types of feedback \cite{keuning.2019} \cite{paiva.2022}. The use of LLMs has opened up new possibilities for the automated creation of teaching materials and the analysis of student work, such as the generation of feedback \cite{prather.2023}.

Hellas et al. \cite{hellas.2023} compared Codex and GPT-3.5 in generating LLM responses to student help requests. They found that in 55\% of student help requests, GPT-3.5 identified and mentioned all actual issues (Codex: 13\%). Notably, 99\% of the responses included code, despite the model being prompted not to do so.
In an attempt to generate feedback as students would do, Kiesler et al. used ChatGPT (March 2023: GPT-3) to examine its responses to incorrect student solutions. In their research 79\% of ChatGPT responses contained code and 62\% contained misleading information \cite{kiesler.2023a}.
Aziaz et al.'s research on feedback generation for programming exercises obtained 52\% fully correct and complete feedback with GPT-4 Turbo \cite{azaiz.2024}, surpassing their previous research with GPT-3.5, which obtained 31\% \cite{azaiz.2023}. The feedback they generated almost always contained code. Since it is difficult to avoid code in the generated feedback, Lifton et al. use a second LLM for "code removal", which rewrites the generated answer \cite{liffiton.2023}.
It appears that the models, including GPT-3.5 and GPT-3.5 Turbo, are biased towards providing a complete solution or code, and they do not consistently follow instructions telling them not to do so \cite{liffiton.2023} \cite{roest.2023}.

\begin{figure*}[!tbh]
    \centerline{\includegraphics[width=0.85\textwidth]{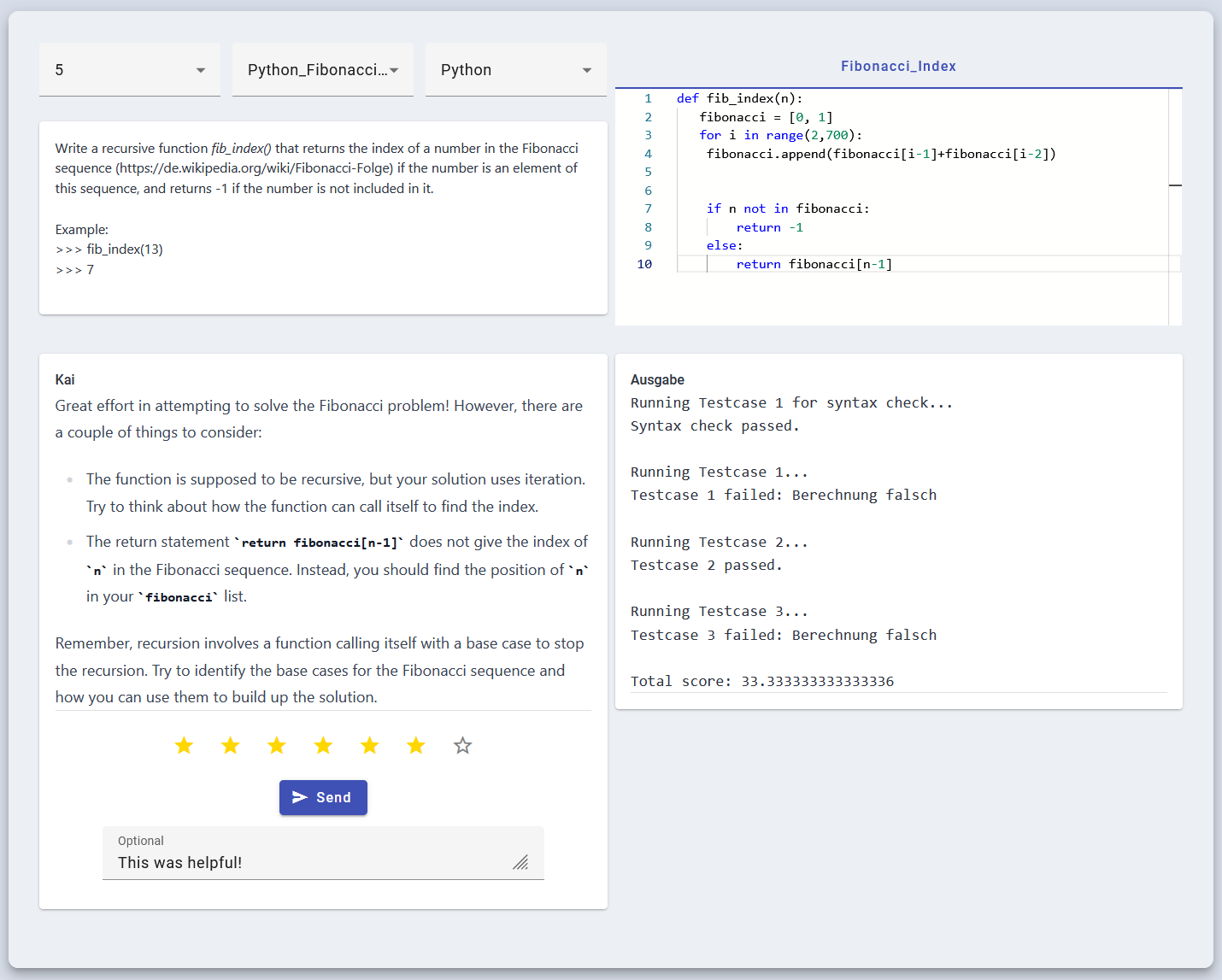}}
    \caption{Tutor Kai user interface}
    \label{fig}
\end{figure*}

It is difficult to compare existing research as the exact model used is not always specified and there are several different models from OpenAI for GPT-3, GPT-3.5, GPT-4 and GPT-4 Turbo. For example, the GPT-4 Turbo is currently available in two different models, "gpt-4-0125-preview" and "gpt-4-1106-preview". Furthermore, the parameters used, such as temperature, which controls the randomness of the response and can therefore have a large effect, are not always reported. 

The input for the LLM (prompt) used also makes a difference \cite{roest.2023} \cite{santos.2023}.  Sometimes only a simple question and the student's code \cite{kiesler.2023a} or additionally the task description \cite{azaiz.2023}\cite{azaiz.2024}\cite{hellas.2023} are used for the prompt. Compiler errors \cite{pankiewicz.2023} and examples of good output (few-shot prompting) \cite{kazemitabaar.2024} can additionally be added for context. Phung et al. show that it can also be helpful to add the result of test cases to the prompt. They also show how GPT-3.5 can be used to simulate a student response to feedback generated by GPT-4. If the generated student response does not pass the task-specific unit tests, this information can be used again to improve the feedback \cite{phung.2023a}.

For the feedback generation in this work, the model GPT-4-0314 (temperature = 0)  with a complex prompt is used to prevent code in the answers. The generated feedback was evaluated not only by experts, but also directly by the students.

\section{Evaluation}
To evaluate the extent to which LLMs are able to provide feedback for programming education, we have developed a web application called Tutor Kai (Fig. \ref{fig}). In Tutor Kai, computer science students can complete weekly tasks for the introductory course "Object-Oriented and Functional Programming" and receive automated feedback generated by an LLM. Students must also rate the feedback they receive on a scale of 1 to 7.

\subsection{Evaluation Setup}
Within Tutor Kai, students select the current week and one of the associated programming tasks. They are then presented with the task and a code editor that may already contain code. Within the code editor (Fig. \ref{fig}: top right), the student solves the task and can execute their solution. The solution is compiled and unit tests are run to verify its correctness. The student receives both error messages from the compiler and the results of the unit tests as text  (Fig. \ref{fig}: bottom right). 

\begin{figure*}[!tbh]
    \centerline{\includegraphics[width=0.8\textwidth]{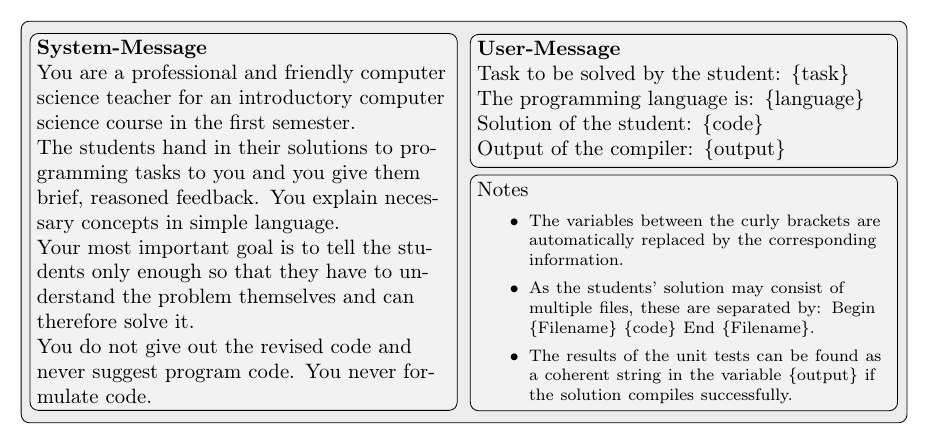}}
    \caption{Prompt (translated from german)}
    \label{fig:Prompttemplate2023}
\end{figure*}

However, the problem with semantic bugs in particular is that while it is communicated that there is a bug, no information is provided about what it is related to. This is where the strength of LLMs for reasoning, coding, and human like text generation comes into play. Students can therefore request feedback after they have executed the program code. For this purpose, the prompt (Fig \ref{fig:Prompttemplate2023}) is sent to the GPT-4 API of OpenAI containing the task, the programming language, the program code, the compiler output and the result of the unit tests as context. We used the first version of GPT-4, which is currently available under the name "gpt-4-0314". The temperature parameter was changed to zero to decrease randomness. The LLM feedback generated in this way must be rated by the student (Fig. \ref{fig}: bottom left).

\subsection{Method}
When evaluating generated feedback, the question arises as to whether it should be viewed atomically or in its chronological sequence in the context of further code submissions per student. The advantage of the second method would be that it is immediately apparent whether the feedback has helped when the student has received more points in a subsequent code submission. However, it cannot be ruled out that students have consulted other external sources to obtain information or have used a different development environment in combination. The second option would therefore only be practical if it could be ensured that only the evaluation setup presented here was used. Such an investigation in a controlled environment is planned for the future.

Based on the methodology of Hellas et al. \cite{hellas.2023}, we therefore analyzed the feedback atomically using the following categories:
\begin{enumerate}
    \item Identification and mention of at least one actual issue
    \item Identification and mention of all issues
    \item Wrong suggestions for improvement
    \item Hallucinated issues
    \item Unnecessary suggestions for improvement
    \item Includes code
\end{enumerate}
Identification and mention of at least one actual issue is present if at least one problem, bug or error that prevents a correct solution is mentioned in the feedback. Likewise, we flagged when all of them were mentioned. Wrong suggestions for improvements are present if the feedback gives a wrong hint or suggests a direction that leads to a wrong solution. Hallucinated issues are issues where non-existent problems, bugs or errors were mentioned in the code submission. Only code snippets that are longer than one line have been marked as code. Individual variables and expressions were not marked, as these are necessary for understanding the feedback. Although these could also be avoided, this would make the feedback longer and more difficult to understand.

To learn more about the students' perspective on how they perceived the feedback, we collected additional data. After they received feedback and before they could submit new code, they were forced to rate the feedback on a scale of one to seven.  Tutor Kai was once offered to students by email for voluntary use.

\section{Results}
In this section, we present the overall aggregated data over the course of the semester and take a closer look at the feedback that was generated for three specific tasks.

\begin{table}[!b]
    \centering
    \begin{tabular}{lp{0.2\textwidth}}
    \textbf{Category} & \textbf{Result}\\ 
    \midrule
    Number of students & 51 \\
    Number of tasks & 26 \\
    Number of code submissions & 3159  \\
    Number of generated feedback & 1684 \\
    Number of student rated feedback & 1243 \\
    Average feedback rating& 5.54 \\
 Average feedback rating (adjusted)&5.05\\
    \end{tabular}
    \caption{Aggregated Data of all tasks and students}
    \label{tab:Datengrundlage}
\end{table}

\begin{table}[!b]
    \centering
    \begin{tabular}{p{0.2\columnwidth}p{0.7\columnwidth}}
    \textbf{Task} & \textbf{Description} \\
    \hline
    Capital-Value & Create a recursive function named `capitalValue()` to calculate the value of a principal amount of 1000 Euros after `n` years at a constant interest rate. \\
    Maximum-Value & Implement the `max()` method in the `Starter` class within the provided code skeleton (Starter.java), which returns the larger of two natural numbers `a` and `b`, or the common value if both are equal. \\
    Sum & Create a recursive function named `summe` that takes two integers `m` and `n` as arguments and returns the sum of all integers between `m` and `n` (inclusive), with the result being 0 if `m` is greater than `n`. \\
    \end{tabular}
    \caption{Task Overview (translated from german)}
    \label{tab:task_overview}
\end{table}

\begin{table*}[tbh]
    \centering
    \begin{tabular}{lllll}
     & \multicolumn{3}{c}{Task} & \\
    \cmidrule(lr){2-4}
     Category & Capital-Value & Maximum-Value & Sum  &All 3 Tasks\\
    \midrule 
    Programming Language & Python & Java & Python  & \\
    Number of code submissions & 175 & 62 & 26  &263\\
    Number of generated feedback & 97 & 25 & 15  &137\\
    Average Feedback Rating & 6.3 & 5.4 & 4.8  &6.0\\
    Identification and mention of  at least one actual issue & 73 (75\%) & 25 (100\%) & 15 (100\%) &113 (87\%)\\
    Identification and mention of all actual issues & 49 (51\%) & 23 (92\%) & 13 (87\%)  &85 (62\%)\\
    Wrong suggestions for improvement & 14 (14\%) & 1 (4\%) & 1 (7\%) &16 (12\%)\\
    Hallucinated issues & 8 (8\%) & 0 (0\%) & 0 (0\%)  &8 (6\%)\\
    Unnecessary suggestion for improvement & 8 (8\%) & 7 (28\%) & 4 (27\%) &19 (14\%)\\
    Includes code & 4 (4\%) & 0 (0\%) & 0 (0\%) &4 (3\%)\\
 \end{tabular}
    \caption{Feedback Evaluation}
    \label{tab:Eval}
\end{table*}

\subsection{Aggregated Data}
An analysis of the aggregated data from all students across the 26 tasks reveals that Tutor Kai was extensively utilized by the participants (Table \ref{tab:Datengrundlage}). Of the 51 students who tried the tool, 25 students used it at least ten times. Twelve students used it more than 100 times.
Overall, the students gave the feedback an average rating of 5.54 on a scale of 1 to 7. There were four students with more than 100 ratings who almost always gave the highest rating for feedback. In interviews it became clear that the forced rating of feedback interfered with task completion and disrupted the flow. Therefore, the highest rating was given immediately in order to continue. After removing these four students from the data set, the average feedback rating drops to 5.05. Another strategy used by the students to avoid the feedback rating was to reload the page. For this reason, the number of feedback ratings given by students is lower than the number of feedback.

\subsection{Individual task evaluation}
The feedback generated on the students' solutions to three tasks (Table \ref{tab:task_overview}) was evaluated using the methodology described. Overall, 87\% of the feedback identified and mentioned at least one actual issue. In 62\% of the feedback, all of them were identified and mentioned  (Table \ref{tab:Eval}).

Hallucinated issues occurred several times in the Capital-Value task for the following reason. The unit tests associated with the task have expected values that are accurate to the twelfth decimal digit. Due to rounding, correct solutions were recognized as incorrect by the unit tests. GPT-4 receives two pieces of divergent information in the prompt template (Fig. \ref{fig:Prompttemplate2023}). On the one hand, the unit tests indicate that the code is incorrect and on the other hand, the correct solution as code. This leads to GPT-4 hallucinating and describing errors in the correct solution that do not exist \cite{Jacobs.2023}.

Unnecessary suggestions for improvement usually occurred when the student's solution was already correct. Using the prompt described (Fig. \ref{fig:Prompttemplate2023}), the model still wants to give feedback. It usually refers to the lack of comments in the solution or suggests alternative ways that could lead to a simpler solution.  A positive correlation between these values and the average feedback rating of the students is not evident.

\section{Discussion}
Despite multiple prompts, the issue of code appearing in the feedback could not be completely avoided. This exemplifies that the behavior of LLMs can neither be fully controlled nor predicted. While the inclusion of code in the feedback may be unproblematic in this case, it could potentially become a more significant issue if LLMs were to be used for grading assignments. 

Compared to the results of Hellas et al. \cite{hellas.2023} and Aziaz et al. \cite{azaiz.2024}, the feedback from Tutor Kai appears to perform better in terms of fully correct and complete feedback at first glance. However, it is unclear whether this is due to the different assignments, the more extensive prompt including the results of the unit tests in Tutor Kai, or the specific LLM used. To attempt to reproduce and compare the results of similar research projects, the exact model and parameters used would need to be known. Even then, it is possible that the providing company may restrict access to the models. 

A problem to be discussed is the deployment and subsequent evaluation of such applications in educational settings. Since the completion of assignments often takes place asynchronously, it cannot be guaranteed that only the evaluated tool was used. To address this, the tools would need to be able to compete with applications like Visual Studio Code in terms of developer experience. During the evaluation of Tutor Kai, the willingness of students to rate the feedback from 1-7 was problematic, even though it only required a single click in the user interface. It remains an open question how this can be improved in the future without negatively impacting the students' workflow too much.

\section{Conclusion}
The evaluation of Tutor Kai demonstrates that the feedback generated by GPT-4 already identifies and mentions most of the issues in the code. Simultaneously, the problem of code appearing in the feedback, which related studies \cite{liffiton.2023} \cite{hellas.2023} have encountered in the past, has been almost entirely resolved. Overall, the students have rated the feedback relatively positively, with an average of 5.05 on a scale from 1 to 7. One issue that was identified is that when students are required to evaluate all feedback, they may seek ways to circumvent this process, potentially distorting the data. 

Furthermore, it has been shown that faulty unit tests in combination with a correct student solution can lead to hallucinated issues. In such cases, errors are addressed that are not present in the student's solution. To avoid this, care must be taken to ensure that GPT-4 does not receive contradictory information.

\section{Outlook}
In the future, the data already collected will also be evaluated with regard to the different types of feedback \cite{narciss.2008} \cite{keuning.2019}. A framework for automating this process will be developed and evaluated. Based on this, faster iterations of models and prompts would then be possible. 

It is expected that future models such as GPT-5 or Llama3, which may be released as early as 2024, will be even more capable of formulating and explaining code and providing feedback. The maximum number of tokens that can be processed by the LLM is also likely to increase. Google's Gemini 1.5 increases this by a factor of eight compared to GPT-4 Turbo, to over one million. This increases the potential for prompting strategies and the use of Retrieval Augmented Generation (RAG) to fill the context window with relevant information about the situational and individual circumstances of students or the content of a lecture. It is therefore already possible to use complete lecture notes or textbooks as input for the LLM when generating feedback. How the increasingly large context windows of LLMs can be filled in a didactically meaningful way in educational contexts needs to be researched more intensively in the future.
\bibliographystyle{IEEEtran}
\bibliography{bib}

\end{document}